\documentclass[journal]{IEEEtran}

\usepackage{algorithmic}
\usepackage{url}
\usepackage{enumerate}
\usepackage{algorithm}
\usepackage{algorithmic}
\usepackage{subfigure}
\usepackage{graphicx}
\usepackage{multirow}
\usepackage{bm}
\usepackage{amsmath,amssymb,amsfonts}
\usepackage{arydshln}
\usepackage{microtype}
\usepackage{microtype}
\usepackage{booktabs}
\usepackage{latexsym}
\usepackage{color}

\begin{document}

\title{Cross-lingual Semantic Role Labeling with Model Transfer}

\author{Hao Fei,
        Meishan Zhang,
        Fei Li
        and
        Donghong Ji
\thanks{This work is supported by the National Natural Science Foundation of China (No.61602160, No.61772378), 
the National Key Research and Development Program of China (No.2017YFC1200500), 
the Research Foundation of Ministry of Education of China (No.18JZD015), and 
the Major Projects of the National Social Science Foundation of China (No.11\&ZD189).
\emph{(Corresponding author: Fei Li).}}
\thanks{Hao Fei and Donghong Ji are with the Department of Key Laboratory of Aerospace Information Security and Trusted Computing, Ministry of Education, School of Cyber Science and Engineering, Wuhan University, Wuhan 430072, China (email: hao.fei@whu.edu.cn and dhji@whu.edu.cn)}
\thanks{Meishan Zhang is with the Department of School of New Media and Communication, Tianjin University, Tianjin 300072, China (email: mason.zms@gmail.com).}
\thanks{Fei Li is with the Department of Computer Science, University of Massachusetts Lowell, Lowell, MA, United States (email: foxlf823@gmail.com).}
}

\maketitle

\begin{abstract}
Prior studies show that cross-lingual semantic role labeling (SRL) can be achieved by model transfer under the help of universal features.
In this paper, we fill the gap of cross-lingual SRL by proposing an end-to-end SRL model that incorporates a variety of universal features and transfer methods.
We study both the bilingual transfer and multi-source transfer, under gold or machine-generated syntactic inputs, pre-trained high-order abstract features, and contextualized multilingual word representations.
Experimental results on the Universal Proposition Bank corpus indicate that performances of the cross-lingual SRL can vary by leveraging different cross-lingual features. 
In addition, whether the features are gold-standard also has an impact on performances.
Precisely, we find that gold syntax features are much more crucial for cross-lingual SRL, compared with the automatically-generated ones.
Moreover, universal dependency structure features are able to give the best help, and both pre-trained high-order features and contextualized word representations can further bring significant improvements.
\end{abstract}


\IEEEpeerreviewmaketitle

\section{Introduction}

\IEEEPARstart{S}{emantic} role labeling (SRL) aims to detect predicate-argument structures from sentences, such as \emph{who did what to whom, when and where}.
Such high-level structures can be used as semantic information for supporting a variety of downstream tasks, including dialog systems, machine reading and translation \cite{shen-lapata-2007-using,liu-gildea-2010-semantic,genest-lapalme-2011-framework,gao-vogel-2011-corpus,wang-etal-2015-machine,KhanSK15}.
Despite considerable research efforts have been paid on SRL, the majority work focuses on the English language, where a large quantity of annotated data can be accessible.
However, the problem of the lack of annotated data prevents the development of SRL studies for those low-resource languages (e.g., Finnish, Portuguese etc.), which consequently makes the cross-lingual SRL much challenging \cite{KozhevnikovT13,aminian-etal-2019-cross}.

Previous work on cross-lingual SRL can be generally divided into two categories: annotation projection and model transfer.
The former is based on large-scale parallel corpora between source and target languages, where the source-side sentences are automatically annotated with SRL tags and then the source annotations are projected to the target-side sentences based on word alignment \cite{KozhevnikovT13,yarowsky-etal-2001-inducing,van-der-plas-etal-2011-scaling,Pado2014Cross,akbik2015generating}.
Annotation projection has received the most attention for cross-lingual SRL.
Consequently, a range of strategies have been proposed for enhancing the SRL performance of the target language,
including improving the projection quality \cite{tiedemann-2015-improving},
joint learning of syntax and semantics information \cite{KozhevnikovT13},
iteratively bootstrapping to reduce the influence of noise target corpus \cite{akbik2015generating},
and joint translation-SRL \cite{aminian-etal-2019-cross}.

On the contrary, the model transfer method builds cross-lingual models based on language-independent or transferable features such as cross-lingual word representations, universal part-of-speech (POS) tags and dependency trees, which can be directly projected into target languages \cite{mcdonald-etal-2013-universal,swayamdipta-etal-2016-greedy,daza-frank-2019-translate}.
One basic assumption under model transfer is the existence of the common features across languages \cite{mcdonald-etal-2013-universal}.
Mulcaire et al. (2018) built a polyglot model by training one shared encoder on multiple languages for cross-lingual SRL \cite{mulcaire-etal-2018-polyglot}.
Daza et al. (2019) proposed an Encoder-Decoder structure model with copying and sharing the parameters across languages for SRL \cite{daza-frank-2019-translate}.
Recently, He et al. (2019) achieved competitive performances by using the multi-lingual contextualized word representation from pre-trained language models \cite{he-etal-2019-syntax}, including the ELMo \cite{PetersNIGCLZ18}, BERT \cite{devlin-etal-2019-bert} and XLM \cite{ConneauL19}.
However, to the best of our knowledge, no prior work of cross-lingual SRL has comprehensively investigated the model transfer method using various cross-lingual features and settings.

In this paper, we conduct elaborate investigation for model-transfer-based cross-lingual SRL, by exploiting universal features (e.g., universal lemmas, POS tags and syntactic dependency tree), pre-trained high-order features and multilingual contextualized word representations.
In addition, we explore the differences between gold and auto-predicted features for cross-lingual SRL under the settings of bilingual transferring and multi-source transferring, respectively.
Last but not least, to enhance the effectiveness of model transfer and reduce the error propagation in the pipeline-style SRL,
we propose an end-to-end SRL model, which is capable of detecting all the possible predicates and their corresponding arguments in one shot.
We also enhance our proposed model with a linguistic-aware parameter generation network (PGN)  \cite{platanios-etal-2018-contextual} for better adaptability on the cross-lingual transfer.

In summary, the contributions of this paper for the cross-lingual SRL are listed as below:
\begin{enumerate}
    
\item We compare various universal features for cross-lingual SRL, including lemmas, POS tags and gold or automatically-generated syntactic dependency trees, respectively.
    We also investigate the effectiveness of pre-trained high-order features,
    and contextualized multilingual word representations that are obtained from certain state—of-the-art pre-trained language encoders such as ELMo, BERT and XLM.
    
    \item We conduct the cross-lingual SRL under both bilingual and multi-source transferring settings to explore the influences of language preferences for model transfer.

    \item We exploit an end-to-end SRL model, jointly performing predicate detection and argument role labeling.
    We reinforce the SRL model with a linguistic-aware PGN encoder.
    We perform empirical comparisons for both the end-to-end SRL model and standalone argument role labeling model.

\end{enumerate}

We perform experiments on the Universal Proposition Bank (UPB) corpus (v1.0) \cite{akbik2015generating,akbik2016polyglot}, which contains annotated SRL data for seven languages.
First, we verify the effectiveness of our method on the single-source bilingual transfer setting,
where the English language is adopted as the source language and another language is selected as the target language.
Results show that the performances of bilingual SRL transfer can vary by using different cross-lingual features, as well as using gold or auto-features.
Furthermore, we evaluate our method on the multi-source transfer setting via our linguistic-aware \emph{PGN} model,
where one language is selected as the target language and all the remaining ones are used as the sources.
We attain better results, but similar tendencies compared with the ones in the single-source transfer experiments,
We conduct detailed analyses for all the above experiments to understand the contributions of cross-lingual features for SRL.
To conclude, first, the gold syntax features are much more crucial for cross-lingual SRL, compared with the automatically-generated features.
Second, the universal dependency structure features give the best help, among all the syntax features. When combining all the universal syntax inputs, our model is able to achieve the best results.
Third, pre-trained high-order features can provide very useful semantic information for supporting the model transfer,
and contextualized word representations are also helpful to improve our model significantly.

\section{Related Work}

\subsection{Semantic Role Labeling}

A line of research efforts has been made for semantic role labeling.
The traditional approaches on SRL are mostly accomplished by designing hand-crafted features and employing statistical machine learning methods \cite{pradhan-etal-2005-semantic-role,PunyakanokRY08,zhao-etal-2009-multilingual-dependency}.
Later, an increasing of attention is drawn for neural networks, consistently bringing improvements for SRL \cite{he-etal-2017-deep,strubell-etal-2018-linguistically,fitzgerald-etal-2015-semantic,roth-lapata-2016-neural,marcheggiani-titov-2017-encoding,he-etal-2018-syntax}.
On the other hand, researchers found that the syntax features are extraordinary effective for the SRL, since intuitively the semantic role structure shares much connections with the underlying syntactic structure.
For example, FitzGerald et at., (2015) integrated lexicalized features into neural models.
% Roth and Lapata et al., (2016) embedded syntactic dependency trees between predicates and arguments.
And Marcheggiani et al., (2017) used the graph convolutional network encoding dependency path.
Strubell et al., (2018) incorporated syntax information into neural model with multi-task learning.

Recently, the end-to-end SRL model has been introduced by He et al., (2018) \cite{he-etal-2018-jointly}, for bringing task improvements with more graceful prediction.
They jointly modeled the predicates and arguments extraction as triplet prediction, effectively reducing the error propagation of the traditional pipeline procedures.
This was followed by the other kinds of reinforced variants for joint SRL model \cite{LiHZZZZZ19}.
In this paper, we consider taking advances of the end-to-end SRL model, while boosting the strength by exploiting a linguistic-aware PGN module for improving cross-lingual SRL transfer.

\vspace{-8pt}
\subsection{Cross-lingual Transfer}

There has been extensive work for cross-lingual transfer learning in natural language processing (NLP) community \cite{KozhevnikovT13,aminian-etal-2019-cross,van-der-plas-etal-2011-scaling,Pado2014Cross,chen-etal-2019-multi-source,RasooliC15,TiedemannA16,0002ZWCMG18,ChenSACW18}.
Model transferring and annotation projection are two mainstream categories for the goal.
The first category aims to build a model based on the source language corpus,
and then adapt it to the target languages \cite{yarowsky-etal-2001-inducing,tiedemann-2015-improving}.
The second category attempts to produce a set of automatic training instances for the target language by a source language model with parallel sentences,
and then train a target model on the dataset \cite{mcdonald-etal-2013-universal,swayamdipta-etal-2016-greedy,daza-frank-2019-translate,mulcaire-etal-2018-polyglot,bjorkelund-etal-2009-multilingual,lei-etal-2015-high}.

For cross-lingual SRL, annotation projection has received the most attention \cite{KozhevnikovT13,aminian-etal-2019-cross,Pado2014Cross,akbik2015generating,tiedemann-2015-improving}.
In contrast, limited efforts have paid for model transfer, where most of them focus on the incorporation of singleton syntax, e.g., universal dependency trees or universal POS tags \cite{daza-frank-2019-translate,mulcaire-etal-2018-polyglot,bjorkelund-etal-2009-multilingual}.
On the other hand, the problematic availability of the gold syntax annotations could prevent the cross-lingual methods for common texts.
In this paper, we conduct investigations of various syntax features, under both auto-predicted ones and gold ones, respectively.
Besides, we explore the pre-trained high-order features, and contextualized word representation from the recent state-of-the-art language models, e.g., ELMo, BERT and XLM, which have been shown to give exceptional task enhancements \cite{he-etal-2019-syntax}.
This work to some extent relates to the recent work of Fei et al. (2020) \cite{Fei06295}, where the authors propose to construct a high-quality pseudo training data at target side for enhancing the cross-lingual SRL.
But in this work, we aim to investigate the cross-lingual features via model transfer method, without changing the benchmark dataset.

\section{Cross-lingual Features}

In this section, we give a full picture on cross-lingual features that we will use for the cross-lingual SRL, as well as their encoding methods, , including cross-lingual word form representation, universal part-of-speech (POS) tags, universal dependency structures,
and the pre-trained high-order features.
We denote them as $\bm{r}^{w}, \bm{r}^{POS}$ $\bm{h}^{tree}$ and $\bm{h}^{ho}$, respectively.
We will use the concatenation of these features as the final input representation for the SRL model:
\begin{equation}
\setlength\abovedisplayskip{2pt}
\setlength\belowdisplayskip{2pt}
    \bm{x}_{i} = \bm{r}^{w}_{i} \oplus \bm{r}^{pos}_{i} \oplus \bm{h}^{tree}_{i} \oplus \bm{h}^{ho}_{i}.
\end{equation}

\subsection{Word Representation}

Given an input sentence $s$, we aim to obtain the word-level representation $\bm{r}^{w}_{i}$ for $i$-th element.
In this study, we consider four types of multilingual word form representations:
(1) word embedding,
(2) ELMo representation,
(3) BERT representation,
and (4) XLM representation.

\textbf{Word embedding} can be obtained from a feed-forward bag-of-word language model, e.g., skip-gram, Cbow \cite{Mikolov1301}, unsupervised pre-training on certain corpus.
Extending the monolingual word embedding to the multilingual one is the most straightforward way for reaching the cross-lingual goal.
In practice, based on the monolingual word embeddings in differing languages, we can exploit aligning tool \cite{museLampleCRDJ18} to project the monolingual ones into a universal space (e.g., English as the pivot language).
Besides, lemma token of each predicate word is exploited as the replacement of raw word for enhancing the information capacity.

\textbf{ELMo} is a language model built upon stacked bidirectional LSTM, which has an exceptional capacity on capturing complex context-sensitive language features in various linguistic contexts \cite{PetersNIGCLZ18}.
Particularly, ELMo uses character-based embedding via Convolutional Neural Network (CNN) to relieve the out-of-vocabulary (OOV) issue, and adopts three-layer LSTM architecture.
Based on the vanilla forward LSTM, ELMo adds a backward-LSTM, predicting words backward, at the same time.
Technically, there can be two manners to achieve the multilingual ELMo word representations,
first, by training the ELMo model based on language-blended datasets \cite{mulcaire-etal-2019-low},
and second, aligning monolingual ELMo representations into unified space \cite{schuster-etal-2019-cross}.

\textbf{BERT} is the state-of-the-art contextualized language model built based on multi-layer Transformer encoders \cite{VaswaniSPUJGKP17}.
BERT employs the unsupervised word mask language modeling (MLM) technology under bi-direction, allowing capturing broader context and making the learned features of language most informative. 
Likewise, the monolingual word representation can be obtained by training the BERT on monolingual corpus.
We can pretrain the BERT on a language mixed corpora to obtain the multilingual word representations.
Besides, we use the averaged vector of the inner-word piece representations from BERT outputs as the full word representation.

\textbf{XLM} is a cross-lingual language model built based on the BERT architecture.
Compared with vanilla BERT, XLM provides much enhanced word mask technology for better training on the multilingual texts, including
1) unsupervised causal language modeling where the words read towards mono-direction, 
and 2) supervised translation language modeling, where the parallel sentence pairs are taken as inputs and perform MLM for enhancing the cross-lingual capability.

\subsection{{Universal POS Tag}}

POS tag is one of the essential linguistic features, which indicates the grammatical role of elements constituent in sentence.
POS tag can provide each word with extended context and the relationship with adjacent component.
% Taking as the example in Fig. \ref{dependency}, the word `\emph{have}' servers as a role of \emph{verbs} (\emph{VB}) in the sentence, indicating the \emph{predicate} role in the semantic structure.
In cross-lingual transfer setting, the universal POS tags are employed, where the consistent annotations are adopted across all languages.
Given an input sentence $s$ with a sequence of POS tags $\{pos_1, \cdots, pos_n\}$, we can obtain the vectorial representations $\bm{r}^{pos}_{i}$ of $i$-th POS tag via a look-up table $\bm{E}^{pos}$.
POS tags can come from gold sources (e.g., in UPB), while we also consider exploiting the automatically predicated POS tags for cross-lingual SRL.
Technically, the auto-POS tags can be generated from POS tagger, which is generally a sequence labeling model, e.g., bi-directional LSTM with Conditional Random Field model (BiLSTM-CRF):
\begin{equation}
\setlength\abovedisplayskip{2pt}
\setlength\belowdisplayskip{2pt}
\text{\em pos}_{1} \cdots \text{\em pos}_{n} = \text{BiLSTM-CRF}(\bm{x}_{1} \cdots \bm{x}_{n}) \,.
\end{equation}

\vspace{-10pt}
\subsection{Dependency Tree}

Rich body of previous work consistently verifies the usefulness of dependency syntax structure for SRL \cite{he-etal-2018-syntax},
because the semantic structure shares much underlying similarity to the syntactic one.
% , as exemplified by the example in Fig. \ref{dependency} (a) and (c).
Intuitively, such intrinsic connections can be leveraged effectively from syntactic dependency tree for facilitating the SRL task.
In addition to the gold dependency tree annotations, we also exploit the automatically predicated dependency syntax, via the state-of-the-art dependency parser, i.e., the Biaffine parser \cite{DozatM17}.
In dependency syntax tree, all the nodes are input words and directed with dependent edges.
Technically, we consider modeling the dependency structure by employing two typical tree encoders: TreeLSTM \cite{miwa-bansal-2016-end}, and graph convolutional network (GCN) \cite{marcheggiani-titov-2017-encoding}, both under bidirectional setting for ensuring sufficient information flow.

\textbf{TreeLSTM} encodes each node $j$ with its corresponding head word as input $x_j$.
Consider the bottom-up one:
\setlength\abovedisplayskip{2pt}
\setlength\belowdisplayskip{2pt}
\begin{align} \label{dependency TreeLSTM}
\overline{h}^{\uparrow}_j &=   \begin{matrix}\sum_{k \in C(j)} \end{matrix} h^{\uparrow}_k \\
i_j &= \sigma(\bm{W}^{(i)} x_j + \bm{U}^{(i)} \overline{h}^{\uparrow}_j + b^{(i)}) \\
f_{jk} &= \sigma(\bm{W}^{(f)} x_j + \bm{U}^{(f)} \overline{h}^{\uparrow}_k + b^{(f)}) \\
o_j &= \sigma(\bm{W}^{(o)} x_j + \bm{U}^{(o)} \overline{h}^{\uparrow}_j + b^{(o)}) \\
u_j &= \text{tanh}(\bm{W}^{(u)} x_j + \bm{U}^{(u)} \overline{h}^{\uparrow}_j + b^{(u)}) \\
c_j &= i_j \odot u_j +  \begin{matrix}\sum_{k \in C(j)}\end{matrix} f_{jk}  \odot c_k \\
h^{\uparrow}_j &= o_j \odot \text{tanh}(c_j)
\end{align}
where
$C(j)$ is the set of child nodes of $j$.
$h_j, i_j, o_j, c_j$ are the hidden state, input gate, output gate and memory cell of node $j$, respectively.
$f_{jk}$ is forget gate for each child $k$ of $j$.
Similarly, we can compute the top-down TreeLSTM as $h^{\downarrow}$.
We use the concatenation of two direction tree representations for each node: $h^{tree}_j = [ h^{\uparrow}_j ; h^{\downarrow}_j]$.

\textbf{GCN} is more computationally efficient, compared with TreeLSTM, performing tree propagation for each node in parallel with O(1) complexity.
Consider the constructed dependency graph
${G}^{\text{dep}} = (\mathcal{V}^{\text{dep}}, {E}^{\text{dep}})$,
where $\mathcal{V}^{\text{dep}}$ are sets of nodes, and ${E}^{\text{dep}}$ are sets of bidirectional edges between heads and dependents, respectively.
GCN can be viewed as a hierarchical node encoder, which represents the node j at the $k$-th layer encodes node $j$:
\begin{align}
\label{gcn gate} g_i^{k} &= \sigma(\bm{W}_i^{k} h_i^{k} + b_i^{k})  \\
\label{dependency GCN} h_j^k  &= \text{ReLU}(  \begin{matrix}\sum_{i \in \mathcal{N}(j)} \end{matrix}  x_i^{k} \odot g_i^{k}  )
\end{align}
where $\mathcal{N}(j)$ are neighbor nodes of $j$ in ${G}^{\text{dep}}$.
ReLU is a non-linear activation function.
We take the final layer's output as the final tree representation $h^{tree}_j$.

\begin{figure}[!t]
\centering
\includegraphics[width=.96\columnwidth]{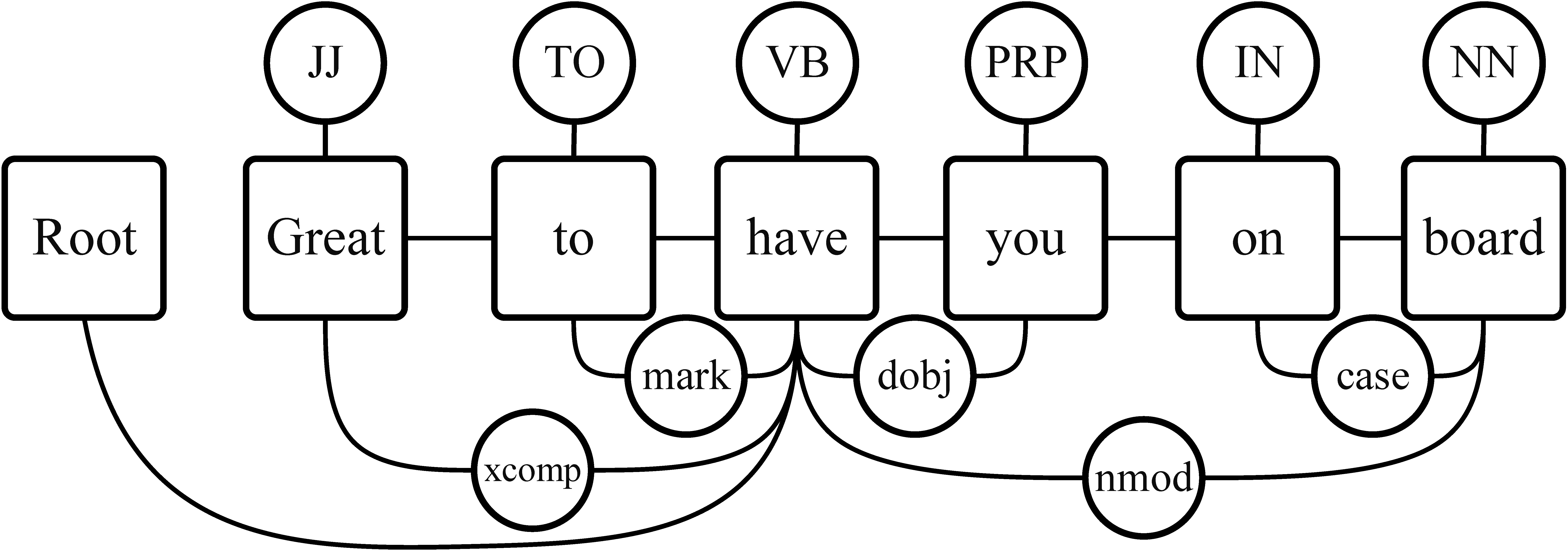}
\caption{
An example to construct a graph for pre-training high-order features.
The sentential words (including the virtual \emph{Root} node) are the main nodes (in square boxes).
The POS tags and dependency labels are the syntactic nodes (in circles).
}
\label{graph-training}
\end{figure}

\subsection{Pre-trained High-order Features}

In addition to the above surface syntax features, we next consider making use of an abstract semantic feature.
We set the target to fully capture the underlying higher order connections between the possible arguments and predicates as much as possible.
Considering the close relevance between syntactic structure and the SRL structure, we thus can enhance the representation of each word in the sentence.
Technically, we can reach this by modeling the syntactic structure as well as the sentential words into one united graph ${G}^{\text{ho}} = (\mathcal{V}^{\text{ho}}, {E}^{\text{ho}})$, where the $\mathcal{V}^{\text{ho}}$ are the nodes including the word tokens (as main nodes) or syntactic labels (i.e., POS tags and dependency labels, as syntactic nodes), and the ${E}^{\text{ho}}$ are the factual edges between each pair of nodes.
Fig. \ref{graph-training} illustrates the relational connections of ${G}^{\text{ho}}$.

We then use a graph neural network (GNN) \cite{KipfW17} to aggregate the information within the graph.
Intuitively, GNN performs propagation over the graph in a parallel manner, by which we can expect the model to capture the abstract higher order semantics for SRL task.
We can pre-train the GNN so that it maps nodes appearing in their structural contexts into the corresponding vectorial embeddings.
Inspired by the work of Hu et al. (2020) \cite{HuLGZLPL20}, we consider the node-level self-supervised pre-training for both the nodes and edges.
Specifically, for \textbf{node prediction}, we randomly mask input nodes (i.e., by replacing them with special mask token), and then make the GNN to predict the labels of the masked nodes (i.e., main nodes or syntactic nodes) based on their neighbor contexts.

Similarly, for \textbf{edge prediction}, we randomly make inference whether there is an arc between each pair of nodes based on their node embeddings.
And the prediction is formulated as a binary classification, i.e., is an arc or is not an arc.
We also employ the negative sampling to accelerate the pre-training.
After pre-training, we can make use of the final representations ($\bm{h}_t^{ho}$) of the main node (words) as enhanced word feature representation.
To adapt the features into cross-lingual scenario, in our practice we conduct the pre-training based on the language-blended corpora.

\section{End-to-end SRL Model}

\subsection{Task Modeling}

Traditional SRL first detects predicates, then followed by the argument extraction, which is a pipeline procedure \cite{TanWXCS18,LiHZZZZZ19}.
In this work, we consider formulating the SRL task in end-to-end scheme.
Concretely, we detect the possible predicates as well as their argument text tokens and the corresponding role labels.
With this respect, the end-to-end formulism enjoys the advantages in two folds.
It is much elegant than the traditional pipeline counterpart, extracting all the possible predicates and the arguments in one shot, meanwhile enjoying the advantages of avoiding the additional noises.

The traditional pipeline models take as input the additional predicate indicator representations, to explicitly indicate the specific predicate position in the sentence.
In contrast, the end-to-end model does need to take predicate indicator information.
Technically, given a sentence $s=\{w_1, w_2, \cdots, w_n \}$ ($n$ is the length of sentence), the SRL model predicts a set of labeled predicate-argument-role triplet $Y \subseteq \mathcal{P} \times \mathcal{A} \times \mathcal{L}$, where $\mathcal{P}=\{w_1, w_2, \cdots, w_n \}$ is the set of all possible predicate tokens, $\mathcal{A}=\{w_1, w_2, \cdots, w_n \}$ contains all the candidate argument tokens\footnote{In this paper, we experiment on the UPB corpus, where the argument text is only annotated on the syntactic head, which also is named the dependency-based SRL \cite{LiHZZZZZ19}. 
Thus, we only need to detect the head token.}, and $\mathcal{L}$ is the space of the semantic role labels, including a null label $\epsilon$ indicating no relation between a candidate predicate-argument pair.
Finally, we expect the model to output all the non-empty relational triplets $\{(p,a,l) \in Y | l \neq \epsilon\}$.

\begin{figure}[!t]
\centering
\includegraphics[width=1.0\columnwidth]{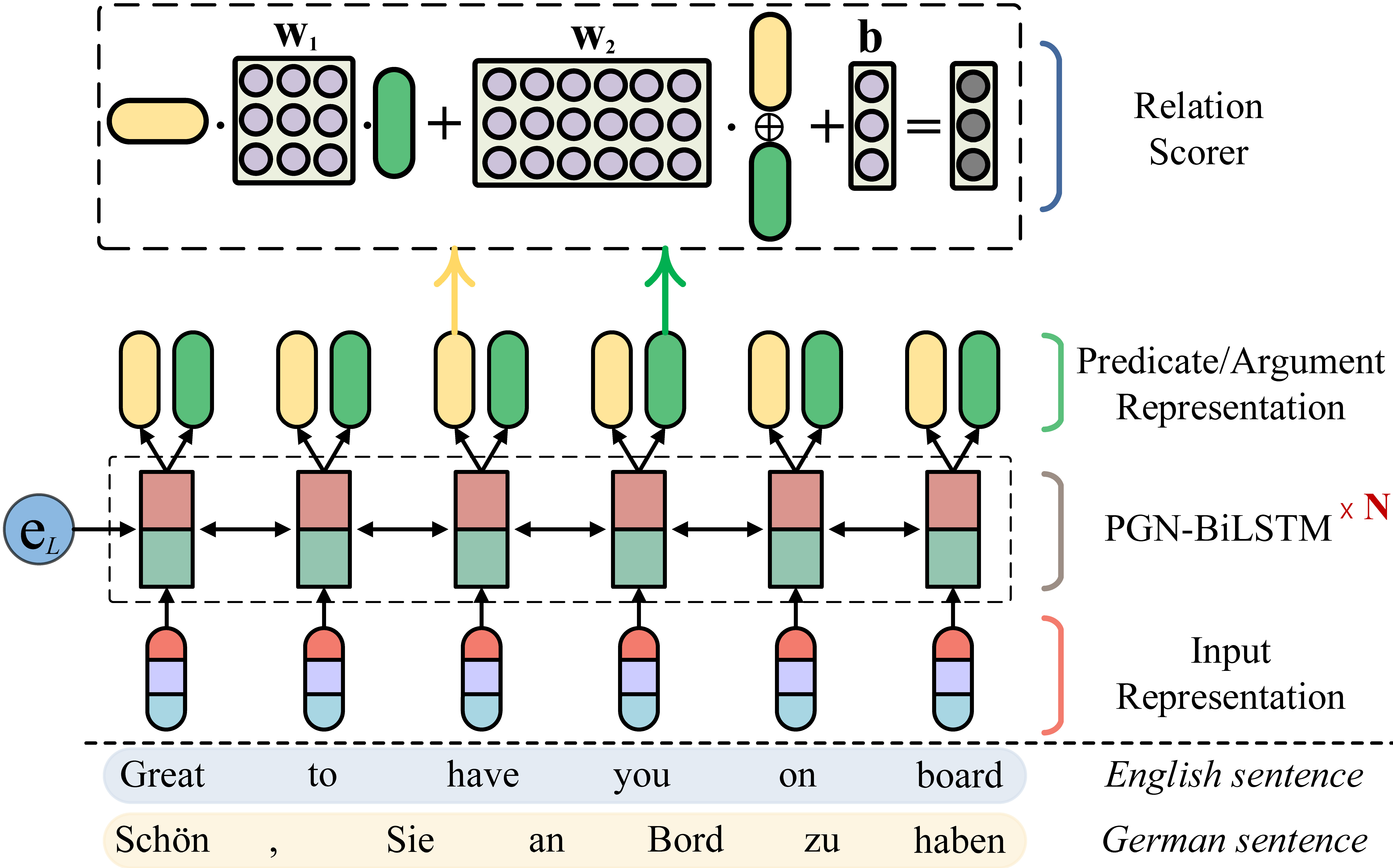}
\caption{
The overall framework of our end-to-end SRL model.
}
\label{framework}
\end{figure}

We exploit a graph-based model, with a similar architecture of He et al. (2018) \cite{he-etal-2018-jointly}, as illustrated in Fig. \ref{framework}.
The model first takes the universal input feature representations, and then the encoder contextualizes the inputs into hidden representation.
The most difference from the one in He et al. (2018) is that our model is equipped with a parameter generation network (PGN), which can help the BiLSTM encoder to capture the linguistic-aware features in cross-lingual SRL.
Afterwards, the model exhaustively enumerates all possible candidate mention representation and builds the element-graph.
Then, the predicate-argument-role triplets will be dynamically decided by the upper relation scorer.
Once the relation between a predicate-argument pair is confirmed valid, the model outputs the triplet.

\subsection{Encoder}

We employ the PGN-BiLSTM \cite{platanios-etal-2018-contextual,jia-etal-2019-cross,Fei06295} to better model the language-aware characteristics.
We use $N$-layer PGN-BiLSTM to contextualize the input representation $\bm{x}$.
\begin{equation}
\begin{split}
\bm{h}_{1} \cdots \bm{h}_{n} & = \text{PGN-BiLSTM}(\bm{x}_{1} \cdots \bm{x}_{n}, \bm{e}_{\mathcal{L}}), \\
    &= \text{BiLSTM}_{\bm{V}_{\mathcal{L}}} (\bm{x}_{1} \cdots \bm{x}_{n}),
\end{split}
\end{equation}
where $\bm{V}_{\mathcal{L}}$ are the flattened vector of all the parameters of the base BiLSTM cell for language $L$, which is dynamically selected by the PGN module:
\begin{equation}
    \bm{V}_{\mathcal{L}} = \bm{W}_{\text{PGN}} \times \bm{e}_{\mathcal{L}},
\end{equation}
where $\bm{e}_{\mathcal{L}}$ is an extra language id label representation for language $L$.
$\bm{W}_{\text{PGN}}$ denotes the parameters of vanilla BiLSTM part in the PGN-BiLSTM, i.e., input, forget, output gates and the cell modules.

\vspace{-8pt}
\subsection{Predicate and Argument Representation}

Based on the token-level hidden representation, we use the relation scorer to determine the predicate-argument pairs.
However, directly using the token representation as the shared common representation can be problematic for distinguish the predicate and argument, since intuitively they play total differing roles in the semantic structure.
Therefore, we apply feedforward layers upon the token level encoder for separating the predicate and argument representation.
\setlength\abovedisplayskip{2pt}
\setlength\belowdisplayskip{2pt}
\begin{align}
    \bm{r}^{p}_i &= \text{FFN}^{p}(\bm{h}_{i}), \\
    \bm{r}^{a}_i &= \text{FFN}^{a}(\bm{h}_{i}),
\end{align}
where $\text{FFN}(\cdot)$ is the corresponding feedforward layer.
The $\bm{r}^{p}, \bm{r}^{a}$ will carry enough unique information to identify the predicate and argument of word $w_i$, respectively.

\vspace{-8pt}
\subsection{Relation Scorer}

Next, we expect the model to decide whether a pair of candidate predicate-argument $\bm{r}^{prd}_i, \bm{r}^{arg}_j$ is valid.
We reach the goal by measuring the corresponding scores via the Biaffine attention, which is a linear transformation structure, first introduced for parsing dependency tree \cite{DozatM17}:
\begin{equation}
\bm{\Phi}(r^{p}_i, r^{a}_j) = \bm{r}^{p}_i \cdot \bm{W}_{1} \cdot \bm{r}^{a}_j  +  \bm{W}_{2} \cdot [\bm{r}^{p}_i;\bm{r}^{a}_j] + \bm{b},
\end{equation}
where $\bm{W}_{1}, \bm{W}_{2}$ are the parameters.

\vspace{-8pt}
\subsection{Training}

During the training, we expect the model to optimize the probability  $P_{\theta}(\hat{y}|s)$ of the triplet $y_{(p, a, l)}$, given a sentence $s$.
Specifically,
\begin{equation}
\begin{aligned}
\label{Learning}
P_{\theta}(y|s) &= \prod_{p \in \mathcal{P}, a \in \mathcal{A}, l \in \mathcal{L}} P_{\theta}(y_{(p, a, l)}|s), \\
                         &= \prod_{p \in \mathcal{P}, a \in \mathcal{A}, l \in \mathcal{L}} \frac{\bm{\Phi}(p, a, l)}{\sum \bm{\Phi}(\hat{p}, \hat{a}, \hat{l})},
\end{aligned}
\end{equation}
where $\theta$ is the parameters of the overall model, $\bm{\Phi}(p, a, l)$ represents the total score:
\begin{equation}
\bm{\Phi}(p, a, l) = \bm{\Phi}(r^{p}_i, r^{a}_j) + \text{FFN}(l_{r^{a}_j}),
\end{equation}
where the additional feedforward layer $\text{FFN}(l_{r^{a}_j})$ measures the scores of the role label $l$ for argument $r^{a}_j$.

Finally, the objective is to minimize the negative log likelihood of the golden structure:
\begin{equation}
\mathcal{L}oss = - log P_{\theta}(y|s).
\end{equation}
Note that the score $P_{\theta}(y_{(p, a, l)}|s)$ of null relation label $\epsilon$ is assigned to 0, which represents an invalid triplet.

As our model exhaustively enumerates all possible triplets, there comes an expensive computation, i.e., $N = \frac{n^2 K}{2}$ ($K$ is the total number of role labels).
Therefore, we define beams to store predicate and argument candidates, with the beam sizes limited in $\alpha_p \cdot n$ and $\alpha_a \cdot n$, respectively.
All candidate predicates and arguments are ranked by their unary score $\bm{\Phi}_p, \bm{\Phi}_a$,
\begin{align}
    \bm{\Phi}_p &= \bm{W_p} \text{FFN}^p(h_i), \\
    \bm{\Phi}_a &= \bm{W_a} \text{FFN}^a(h_i),
\end{align}
and only those within the corresponding beams will be cached.

\vspace{-8pt}
\subsection{Inference}

In the inference stage, the well-trained model will output all possible triplets $(p, a, l)$.
In particular, after the measurement by relation scorer, a softmax classifier is used to predict role label $l$
for the valid predicate-argument $r^p_i, r^a_j$ candidate:
\begin{equation}
\label{r-softmax}
l = \text{softmax}( r^a_j)
\end{equation}
If the semantic label is not null $\epsilon$, the label $l$, and the candidate predicate $r^p_i$, argument $r^a_j$ will be output according to their start and end index.

\begin{table}[!t]
\begin{center}
  \caption{Statistics of the UPB corpus,
  where \texttt{Fam.} indicates the language family,
  IE.Ge refers to the Indo-European Germanic, IE.Ro refers to the Indo-European Romance.
  }
  \label{Dataset of UPB}
\resizebox{1.0\columnwidth}{!}{
  \begin{tabular}{ccrrrrr}
\hline
 \texttt{Fam.}  & \texttt{Lan.}  & \texttt{Train}  & \texttt{Dev} &  \texttt{Test}  & \texttt{Prd.}& \texttt{Arg.} \\
\cmidrule(r){1-2}\cmidrule(r){3-5}\cmidrule(r){6-7}
\multirow{2}{*}{IE.Ge} & EN &10,907&	1,631&	1,633&	41,359&	100,170\\
 & DE & 14,118&	799&	977&	23,256&	58,319\\
\cmidrule(r){1-2}\cmidrule(r){3-5}\cmidrule(r){6-7}
\multirow{4}{*}{IE.Ro} & FR &14,554&	1,596&	298&	26,934&	44,007\\
 & IT &12,837&	489&	489&	26,576&	56,893\\
 & ES &28,492&	3,206&	1,995&	81,318&	177,871\\
 & PT &7,494&	938&	936&	19,782&	41,449\\
\cmidrule(r){1-2}\cmidrule(r){3-5}\cmidrule(r){6-7}
Uralic & FI &12,217&	716&	648&	27,324&	60,502\\
\hline
\end{tabular}
}
\end{center}
\end{table}

\section{Experimental Settings}

\subsection{Data}

Our experiments are based on Universal Proposition Bank (UPB, v1.0) \footnote{\url{https://github.com/System-T/UniversalPropositions}}, which is built upon Universal Dependency Treebank (UDT, v1.4)\footnote{\url{https://lindat.mff.cuni.cz/repository/xmlui/handle/11234/1-1827}} and Proposition Bank (PB, v3.0)\footnote{\url{http://propbank.github.io/}}.
In UPB, consistent dependency-based universal SRL annotations are constructed across all languages.
We additionally assemble the English SRL dataset based on the English EWT subset from the UDT v1.4 and the English corpus in PB v3.0.
Finally, we choose a total of seven languages,
including English (EN) and German (DE) of the IE.German family, French (FR), Italian (IT), Spanish (ES) and Portuguese (PT) of the IE.Romance family, and Finnish (FI) of the Uralic family.
Each subset in different languages comes with its own training, developing and testing set.
Table \ref{Dataset of UPB} shows the data statistics in detail.

\vspace{-8pt}
\subsection{Feature Setup}

% We investigate types of multilingual input representations.
For multilingual word embedding (denoted as \textbf{Emb.}), \texttt{MUSE}\footnote{\url{https://github.com/facebookresearch/MUSE}} is used to align all monolingual fastText word embeddings into a universal space \cite{museLampleCRDJ18}.
For ELMo, a blended dataset\footnote{\url{https://lindat.mff.cuni.cz/repository/xmlui/handle/11234/1-1989}} of the seven languages is used to train multilingual ELMo directly.
For BERT model, the official released multilingual BERT (base, cased version)\footnote{\url{https://github.com/google-research/bert}} is used.
For XLM, we use the official model\footnote{\url{https://github.com/facebookresearch/XLM}} that is pre-trained over 100 languages.
We use the GNN model with 350 dimensional hidden units and 5 layers for the high-order feature pre-training, based on the same blended corpus of training the ELMo model.

All dataset in its language contains gold POS tags and the dependency tree annotations.
As we state earlier, to validate the effectiveness of the auto-predicted syntax features, we further annotate these features via external tools.
In particular, the POS tags of the translated sentences are produced by supervised monolingual BiLSTM-CRF POS taggers,
which are trained on the corresponding UDT v1.4 datasets, respectively, with averaged accuracy of 93.7\% in seven languages.
For the universal dependency, we use the state-of-the-art Biaffine parser \cite{DozatM17} to obtain the tree annotations.
Being trained on the Penn Treebank \cite{PTB93}, the dependency parser achieves an averaged LAS of 93.4\%, and 95.2\% UAS on WSJ test sets.

\begin{figure}[!t]
\centering
\includegraphics[width=.99\columnwidth]{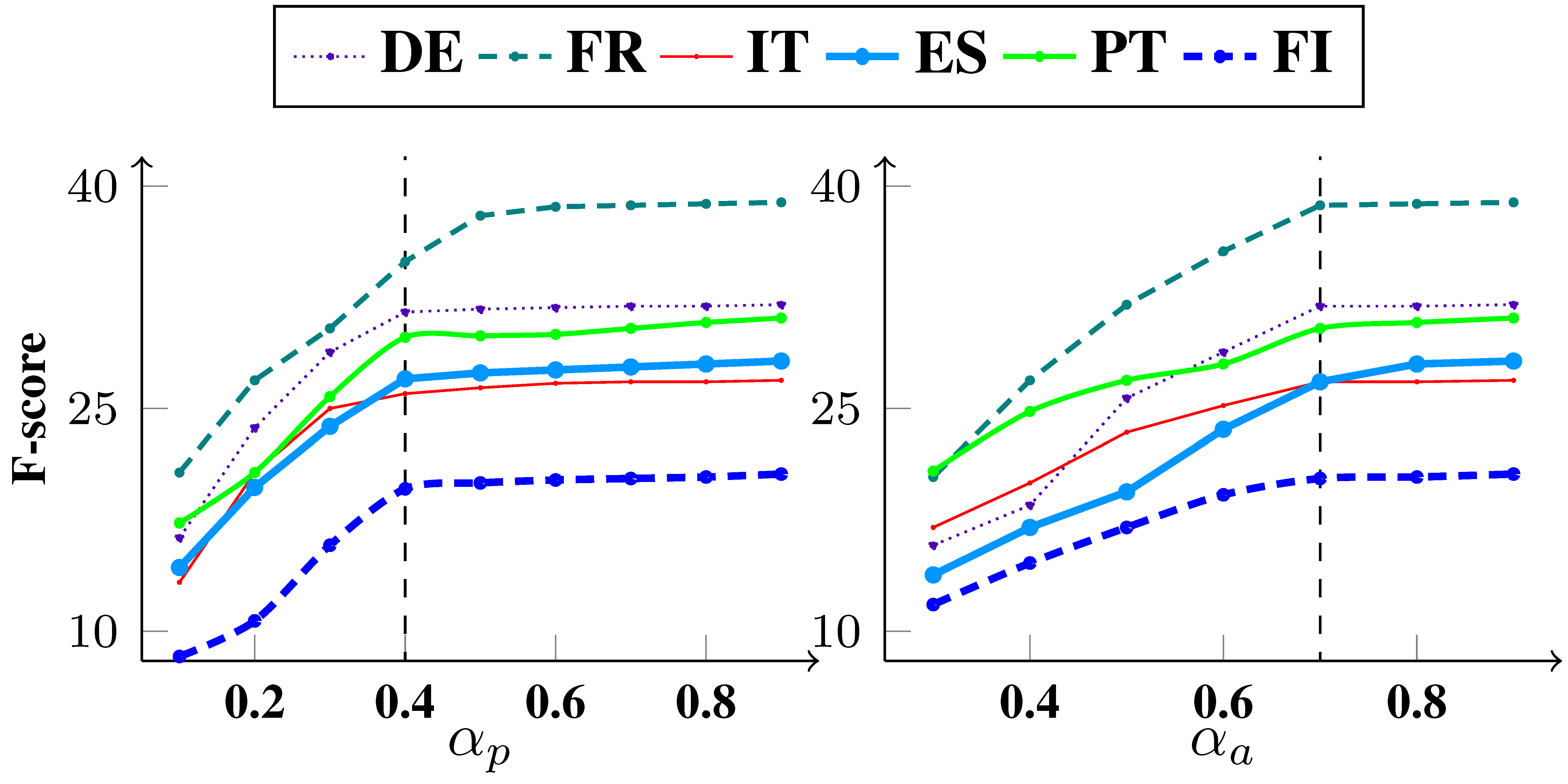}
\caption{
Influences under varying argument and predicate candidate thresholds.
}
\label{threshold}
\end{figure}

\vspace{-8pt}
\subsection{Hyper-parameters}

For the end-to-end SRL models,
the dimension sizes of multilingual word embeddings, ELMo, BERT, XLM and high-order feature representation are 300, 1024, 768, 1024 and 300, respectively.
The POS tag and dependency label embedding sizes are 100, 200 respectively.
The hidden size of BiLSTM (or PGN-BiLSTM) is set to 650, and the hidden sizes for TreeLSTM and GCN are 300.
% Both the TreeLSTM and GCN has 2 layers, and the BiLSTM encoder of our model has 3 layers (N=3).
We exploit online training with a batch size of 30,
and the model parameters are optimized by using the Adam algorithm with an initial rate of 0.001.
The training is performed over the whole training dataset without early-stopping
for 80 iterations on bilingual transfer,
and 300 iterations on multilingual transfer.

\vspace{-8pt}
\subsection{Baselines and Evaluation Metrics}
We denote our PGN-BiLSTM based end-to-end model as \emph{PGN}, while the version where the PGN encoder is replaced with the vanilla BiLSTM is used as our \emph{BASIC} model.
We compare the performances with several joint baselines, including the one by He et al., (2017) \cite{he-etal-2017-deep}, denoted as \emph{He-17}, the model (\emph{He-18}) by He et al., (2018) \cite{he-etal-2018-jointly}, and the model by Strubell et al. (2018) \cite{strubell-etal-2018-linguistically}, denoted as \emph{Strubell-18}.
We also evaluate the results on these same baselines only on the second step of SRL, i.e., argument role labeling, by given the gold predicates.
To compare the effectiveness of our \emph{PGN} model, we additionally employ two state-of-the-art linguistic-aware encoders for multi-lingual transfer, including the \emph{MoE} model by Guo et al. (2018) \cite{guo-etal-2018-multi}, and
the \emph{MAN-MoE} model proposed by Chen et al. (2019) \cite{chen-etal-2019-multi-source}, respectively.
We replace the PGN-BiLSTM in our end-to-end architecture with these encoders.
% , with other modules kept unchanged.

\vspace{-8pt}
\subsection{Development Experiments}

Besides of the regular hyper-parameters, we introduce the predicate and argument beam threshold $\alpha_p, \alpha_a$ for pruning the encoding complexity.
Here we evaluate their influences by the development experiments, based on the transfer from English source to other target languages.
As shown in Fig. \ref{threshold}, for $\alpha_p$, most of the experimental results can converge at 0.4.
And the best performances can be obtained for most of the languages with $\alpha_a$=0.7.
The observation is reasonable, since the predicate numbers in sentences under all languages are less than the numbers of the arguments.
We thus set them with these values universally for all languages on the rest of the experiments.
% Under such pruning, we can maintain a trade-off between efficiency and accuracy.

\begin{table}[!t]
\begin{center}
  \caption{
  Results of cross-lingual transfer from English on end-to-end predicate and argument role labeling.
  }
  \label{Xling-En-end-to-end}
\resizebox{1.0\columnwidth}{!}{
  \begin{tabular}{lccccccc}
\hline
\multicolumn{1}{c}{Feature}  & ~DE~& ~FR~& ~IT~& ~ES~& ~PT~& ~FI~ &\texttt{Avg}\\
\hline \hline
% \hline
\multicolumn{8}{l}{$\bullet$ \textbf{Basic Word Feature (+Emb)}}\\ 
% \hline
 Word &31.5 & 38.3 & 26.8 & 27.0 & 29.8 & 19.6 & 28.7 \\
 \quad +Lemma & 33.1 & 39.6 & 27.7 & 28.8 & 32.0 & 20.3 & 30.3 \\ \hline
\multicolumn{8}{l}{$\bullet$ \textbf{Auto Syntax Feature (+Emb)}}\\ 
% \hline
\rotatebox{0}{POS} & 36.3 & 43.5 & 33.3 & 35.1 & 39.2 & 24.7 & 35.3 \\
\multicolumn{1}{l}{Dependency: }&\multicolumn{6}{l}{}&\multicolumn{1}{l}{}\\
\multicolumn{1}{l}{\quad +TreeLSTM} &	45.9 & 52.5 & 43.9 & 42.5 & 40.2 & 29.3 & 42.4 \\
\multicolumn{1}{l}{\quad +GCN} &	47.0 & 55.1 & 44.2 & 43.2 & 42.3 & 30.4 & 43.7 \\
\cdashline{1-8}
Comb. &  \bf 52.0 &  \bf 60.9 &   \bf 52.1 &   \bf 48.9 &   \bf 48.4 &   \bf 36.5 &   \bf 49.8 \\
\hline
\multicolumn{8}{l}{$\bullet$ \textbf{Gold Syntax Feature (+Emb)}}\\ 
% \hline
\rotatebox{0}{POS} & 38.7 & 45.9 & 36.7 & 38.5 & 42.0 & 27.9 & 38.6 \\
\multicolumn{1}{l}{Dependency: }&\multicolumn{6}{l}{}&\multicolumn{1}{l}{}\\
\multicolumn{1}{l}{\quad +TreeLSTM} &	48.7 & 55.7 & 47.4 & 46.4 & 45.4 & 32.7 & 46.1 \\
\multicolumn{1}{l}{\quad +GCN} &	49.2 & 58.0 & 47.5 & 44.8 & 46.9 & 35.0 & 46.9 \\
\cdashline{1-8}
Comb. &  \bf 	54.7 &   \bf 62.3 &  \bf  56.7 &   \bf 51.5 &   \bf 50.4 &   \bf 38.9 &   \bf 52.4 \\
\hline
\multicolumn{8}{l}{$\bullet$ \textbf{Pre-trained High-order Feature (+Comb. Gold Syn.)}}\\ 
\rotatebox{0}{HO.} & 56.5 & 65.1 & 58.2 & 52.5 & 52.0 & 39.6 & 54.0 \\
\hline
\multicolumn{8}{l}{$\bullet$ \textbf{Contextual Word Representation (+Comb. Gold Syn. +HO.)}}\\ 
% \hline
\rotatebox{0}{ELMo} & 57.2 & 66.4 & 56.9 & 56.4 & 51.5 & 44.2 & 55.4  \\
\rotatebox{0}{BERT} & 58.3 & 68.8 & 57.5 & 58.6 & 53.1 & 45.4 &  56.9 \\
\rotatebox{0}{XLM} & \bf 58.9 &  \bf 69.4 &  \bf 61.1 &  \bf 59.5 &  \bf 53.8 &  \bf 46.5 & \bf 58.2 \\
\hline\hline
\multicolumn{8}{l}{$\bullet$ \textbf{SRL Baseline Model (+Comb. Gold Syn. +HO. +XLM)}}\\
\rotatebox{0}{He-17} & 56.1 & 66.1 & 57.5 & 56.1 & 51.3 & 42.6 & 54.9 \\
\rotatebox{0}{Strubell-18} &57.9 & 68.0 &   \bf 59.9 &   \bf 58.2 & 50.7 & 44.3 & 56.6 \\
\rotatebox{0}{He-18} &  \bf  58.3 &   \bf 68.4 & 58.8 & 58.0 &  \bf  51.9 &  \bf  45.2 &   \bf 57.0 \\
\hline
\end{tabular}
}
\end{center}
\end{table}

\section{Experimental Results}

\subsection{Cross-Lingual Transfer from English}

We first conduct experiments on cross-lingual transfer from the English source to the rest of the other six target languages, respectively,
which has been a typical setting for cross-lingual investigations \cite{wang-etal-2019-cross}.
We first evaluate the performances under end-to-end extracting scheme, and then explore the usefulness of the models for the standalone argument role labeling with pre-identified predicates.
We note that the bilingual transfer only takes as input one source language, and thus we can only employ the \emph{BASIC} model as our backbone.

\begin{table}[!t]
\begin{center}
  \caption{Ablation studies on standalone pre-trained features, including multilingual word representation and high-order features.
  }
  \label{ablation}
\resizebox{0.95\columnwidth}{!}{
  \begin{tabular}{lccccccc}
\hline
\multicolumn{1}{c}{Feature}  & ~DE~& ~FR~& ~IT~& ~ES~& ~PT~& ~FI~ &\texttt{Avg}\\
\hline 
\multicolumn{8}{l}{$\bullet$ \textbf{Contextual Word Representation}}\\ 
\rotatebox{0}{ELMo} & 40.6 & 44.7 & 33.6 & 32.9 & 39.0 & 25.3 & 36.0 \\
\rotatebox{0}{BERT} &  43.3 &  \bf  51.7 & 35.4 & 38.0 & 42.8 & 25.4 & 39.4 \\
\rotatebox{0}{XLM} &   \bf 45.7 & 51.0 &   \bf 38.2 &  \bf  41.2 &   \bf 39.8 &  \bf  28.3 &  \bf  40.7 \\
\hline
\multicolumn{8}{l}{$\bullet$ \textbf{Pre-trained High-order Feature}}\\ 
\rotatebox{0}{HO.} &  \bf 49.8 &  \bf 57.6 &  \bf 47.0 &  \bf 43.5 &  \bf 54.5 &  \bf 39.9 & \bf 48.7 \\
\hline
\end{tabular}
}
\end{center}
\end{table}

\textbf{The end-to-end SRL} results are summarized in Table \ref{Xling-En-end-to-end}.
We list the F-scores by using basic word embedding features, auto-predicted syntax features, gold syntax features, pre-trained high-order feature and contextualized word representation, and different SRL models as well.
First, we find that the lemma can partially play positive role for the performance.
Then, unsurprisingly, the syntax features can improve the transferring, for all the target languages.
Specifically, the POS tags give an average F1 score of 5.0(35.3-30.3)\%, while the contributions from dependency structure by TreeLSTM and GCN are 12.1\% and 13.4\%, respectively, demonstrating the advances from GCN tree encoder.
When combining the POS tags with dependency tree, the performances further increase to 49.8\%, which is a very considerable performance gain.
Such observation can also be found in gold syntax features.
But the gold syntax features universally help to give higher increases, compared with auto-predicted correspondences.
Especially the dependency tree by both the TreeLSTM and GCN encoder, can yield enhanced improvements.
This shows that the gold syntax is much more crucial for the cross-lingual SRL.
% , providing high-quality of features.

Next, based on all the combined gold syntax feature, we demonstrate that the pre-trained high order feature can lead to further impressive performance increases, with average of 1.6\% (54.0-52.4) F1 score.
Further, based on all the above features, we can see that the BERT, ELMo and XLM representations can all help to make additional contributions, where the multilingual BERT is slightly better than the ELMo, while the XLM offers the best utilities.
Finally, we evaluate the performances by different SRL models under the current best setups, i.e., with combined gold syntax and high order features with XLM word representations.
Overall, these end-to-end baselines can yield performances close to our \emph{BASIC} model, with the help of all these cross-lingual features.

\begin{table}[!t]
\begin{center}
  \caption{Results of cross-lingual transfer from English on argument role labeling using gold features. }
  \label{Xling:En:ARL}
\resizebox{1.0\columnwidth}{!}{
  \begin{tabular}{lccccccc}
\hline
\multicolumn{1}{c}{Feature}  & ~DE~& ~FR~& ~IT~& ~ES~& ~PT~& ~FI~ &\texttt{Avg}\\
\hline \hline
\multicolumn{8}{l}{$\bullet$ \textbf{Basic Word Feature (+Emb)}}\\ 
 Word &36.7 & 45.1 & 36.5 & 33.0 & 36.8 & 26.1 & 35.7 \\
 \quad +Lemma & 38.5 & 46.8 & 38.2 & 34.6 & 38.1 & 27.8 & 37.3 \\
 \hline
\multicolumn{8}{l}{$\bullet$ \textbf{Gold Syntax Feature (+Emb)}}\\ 
\rotatebox{0}{POS} & 42.7 & 51.0 & 42.6 & 40.1 & 43.9 & 30.0 & 41.7 \\
% \cdashline{2-8}
\multicolumn{1}{l}{Dependency: }&\multicolumn{6}{l}{}&\multicolumn{1}{l}{}\\
\multicolumn{1}{l}{\quad +TreeLSTM} &	50.2 & 59.8 & 51.8 & 48.3 & 49.0 & 36.6 & 49.3 \\
\multicolumn{1}{l}{\quad +GCN} &	51.6 & 62.3 & 52.7 & 48.9 & 50.3 & 37.8 & 50.6 \\
\cdashline{1-8}
Comb. &	  \bf 56.7 &   \bf 65.5 &   \bf 58.3 &   \bf 55.8 &   \bf 52.2 &   \bf 43.8 &   \bf 55.4 \\
\hline
\multicolumn{8}{l}{$\bullet$ \textbf{Pre-trained High-order Feature (+Comb. Gold Syn.)}}\\ 
\rotatebox{0}{HO.} & 58.9 & 67.8 & 60.2 & 59.2 & 53.7 & 46.6 & 57.7 \\
\hline
\multicolumn{8}{l}{$\bullet$ \textbf{Contextual Word Representation (+Comb. Gold Syn. +HO.)}}\\ 
\rotatebox{0}{ELMo} & 61.5 & 71.4 & 62.9 & 62.3 & 56.4 & 48.4 & 60.5 \\
\rotatebox{0}{BERT} & 62.4 & 70.9 & 63.2 & 62.1 & 58.5 & 49.8 & 61.1 \\
\rotatebox{0}{XLM} & \bf 65.8 & \bf 73.6 &  \bf 65.7 &  \bf 64.1 & \bf 59.2 &  \bf 52.0 &  \bf 63.6 \\
\hline\hline
\multicolumn{8}{l}{$\bullet$ \textbf{SRL Baseline Model (+Comb. Gold Syn. +HO. +XLM)}}\\ 
\rotatebox{0}{He-17} &   \bf 63.8 & 71.6 & 62.6 & 60.4 & 56.6 &   \bf 50.0 & 60.8 \\
\rotatebox{0}{He-18} & 63.1 &   \bf 72.2 &   \bf 63.5 &   \bf 62.2 &  \bf  57.6 & 49.7 &   \bf 61.4 \\
% \hdashline
\hline
\end{tabular}
}
\end{center}
\end{table}

\textbf{Ablation studies on pre-trained features:}
Above we present the performances by stacked features incrementally, examining the best performances on cross-lingual SRL that our model can achieve.
We now take further view on the separate contributions from the pre-trained features including the contextualized multilingual word representations, and the high-order feature, without syntax features combined.
We note that both the contextual word representations and the high-order features can be seen as external sources for offering extended semantic information.
Table \ref{ablation} gives the ablation results.
First, we find that the trends of the usefulness by different contextual language models are kept same with that in Table \ref{Xling-En-end-to-end}, that is, the XLM is most helpful than the ELMo and BERT.
Compared with the syntax features (gold), contextual word representations provide utilities in a level that are slightly lower than that of the syntactic features.
In contrast, the pre-trained multilingual high-order features can support the transfer much prominently, with averaged 48.7\% F1 score, only slightly smaller than the one by combined gold syntax feature (52.4\%).
% The possible reason behind such big contributions largely lies in that, the syntactic knowledge including both the POS and dependency structures can be fully learnt during the self-supervised pre-training, and meanwhile the underlying semantic higher order relationships among the surface words and syntactic structures are sufficiently captured.
Above observations verify the advantages of using such pre-trained semantic information \cite{PetersNIGCLZ18,devlin-etal-2019-bert,HuLGZLPL20}.

\begin{table}[!t]
\begin{center}
  \caption{End-to-end bilingual transfer using different syntax features.
  }
  \label{fine:bilingual1}
\resizebox{0.9\columnwidth}{!}{
\begin{tabular}{cccccccc}
\hline
 Source  & {EN}  & {DE}  & {FR} &  {IT}  & {ES}& {PT} & {FI} \\ 
 \hline\hline
%  \hline
\multicolumn{8}{l}{ $\bullet$ \bf Gold POS Feature } \\
\hline
EN & \multicolumn{1}{|c}{ }  & \multicolumn{1}{c|}{\bf 39.7}  & 45.9  & 37.7  & 38.5  & 42.0  & \bf 27.9  \\
DE & \multicolumn{1}{|c}{\bf 47.6}  & \multicolumn{1}{c|}{ }  & 48.8  & 37.7  & 40.2  & 37.7  & 26.7  \\    \cline{2-7}
FR & 44.2  & 30.0  & \multicolumn{1}{|c}{ }  & 43.3  & 36.0  & \multicolumn{1}{c|}{46.8}  & 26.3  \\
IT & 45.7  & 36.5  & \multicolumn{1}{|c}{\bf 49.4}  &    & 40.5  & \multicolumn{1}{c|}{41.4}  & 25.7  \\
ES & 44.4  & 34.2  & \multicolumn{1}{|c}{44.2}  & \bf 45.7  &    & \multicolumn{1}{c|}{\bf 55.9}  & 25.0  \\
PT & 40.2  & 35.9  & \multicolumn{1}{|c}{43.4}  & 41.0  & \bf 43.5  & \multicolumn{1}{c|}{ }  & 16.8  \\    \cline{4-7}
FI & 31.8  & 26.2  & 48.9  & 30.4  & 39.2  & 34.1  &    \\
\hline
\hline
\multicolumn{8}{l}{ $\bullet$ \bf Gold Dependency Tree Feature (by GCN) } \\
\hline
EN & \multicolumn{1}{|c}{ }  & \multicolumn{1}{c|}{\bf 49.2}  & 58.0  & 47.5  & 44.8  & 46.9  & \bf 35.0  \\
DE & \multicolumn{1}{|c}{\bf 46.3}  & \multicolumn{1}{c|}{ }  & 57.4  & 48.3  & 49.8  & 45.5  & 29.8  \\    \cline{2-7}
FR & 41.0  & 38.3  & \multicolumn{1}{|c}{ }  & 52.4  & 44.2  & \multicolumn{1}{c|}{53.5}  & 29.5  \\
IT & 44.3  & 40.7  & \multicolumn{1}{|c}{58.4}  &    & 45.8  & \multicolumn{1}{c|}{47.2}  & 29.7  \\
ES & 41.4  & 39.2  & \multicolumn{1}{|c}{57.1}  & 52.1  &    & \multicolumn{1}{c|}{\bf 59.8}  & 31.0  \\
PT & 38.6  & 40.2  & \multicolumn{1}{|c}{58.0}  & \bf 52.7  & \bf 50.2  & \multicolumn{1}{c|}{ }  & 18.8  \\    \cline{4-7}
FI & 34.1  & 36.8  & \bf 58.8  & 43.6  & 41.2  & 43.8  &    \\
\hline
\hline
\multicolumn{8}{l}{ $\bullet$ \bf Combined Dold Syntax Features (+XLM) } \\
\hline
EN & \multicolumn{1}{|c}{ }  & \multicolumn{1}{c|}{\bf 56.4}  & 65.9  & 56.1  & 55.2  & 50.8  & \bf 43.0  \\
DE & \multicolumn{1}{|c}{\bf 45.5}  & \multicolumn{1}{c|}{ }  & 64.0  & 59.5  & 57.6  & 47.9  & 39.4  \\    \cline{2-7}
FR & 42.5  & 45.9  & \multicolumn{1}{|c}{ }  & 58.4  & 56.0  & \multicolumn{1}{c|}{58.6}  & 40.7  \\
IT & 40.9  & 50.1  & \multicolumn{1}{|c}{64.7}  &    & 56.5  & \multicolumn{1}{c|}{55.4}  & 38.7  \\
ES & 43.8  & 49.1  & \multicolumn{1}{|c}{\bf 66.9}  & \bf 61.8  &    & \multicolumn{1}{c|}{\bf 62.9}  & 36.4  \\
PT & 40.7  & 50.8  & \multicolumn{1}{|c}{63.5}  & 58.9  & \bf 61.8  & \multicolumn{1}{c|}{ }  & 30.1  \\    \cline{4-7}
FI & 34.5  & 43.3  & 62.9  & 51.6  & 54.7  & 48.9  &    \\
\hline
\end{tabular}
}
\end{center}
\end{table}

\begin{table}[!t]
\begin{center}
  \caption{Results of argument role labeling using XLM representations, gold syntax features and high-order features.
  }
  \label{fine:bilingual4}
\resizebox{0.9\columnwidth}{!}{
\begin{tabular}{cccccccc}
\hline
 Source  & {EN}  & {DE}  & {FR} &  {IT}  & {ES}& {PT} & {FI} \\ \hline\hline
EN & \multicolumn{1}{|c}{ }  & \multicolumn{1}{c|}{\bf 66.8} & 73.6 & 65.7 & 66.1 & 57.2 & \bf 52.0 \\
DE & \multicolumn{1}{|c}{\bf 67.6}  & \multicolumn{1}{c|}{ }  &  72.0 & 66.6 & 65.8 & 53.7 & 49.1  \\    \cline{2-7}
FR & 64.7 & 59.0 & \multicolumn{1}{|c}{ }  & \bf  68.2 & 64.4  & \multicolumn{1}{c|}{60.5}  & 49.3  \\
IT & 62.5 & 64.0  & \multicolumn{1}{|c}{\bf 74.0}  &    & \bf 67.3  & \multicolumn{1}{c|}{61.9}  & 45.9 \\
ES & 65.5 & 65.0  & \multicolumn{1}{|c}{71.7}  & 68.0  &    & \multicolumn{1}{c|}{\bf 62.7}  & 48.6  \\
PT & 62.5 & 66.0  & \multicolumn{1}{|c}{71.9}  & 67.5 & 66.3   & \multicolumn{1}{c|}{ }  &  39.8 \\    \cline{4-7}
FI & 59.5 & 62.6 & 70.8 & 58.1 & 62.2 & 54.6  &    \\
\hline
\end{tabular}
}
\end{center}
\end{table}

\textbf{Argument role labeling:} Table \ref{Xling:En:ARL} shows the performances for the standalone argument role labeling of the second step of SRL.
Overall, the similar trends for different features and settings can be observed as that in end-to-end SRL.
But the scores are universally better, since here the predicates are given as gold signals and the task degraded into a simpler one.
We see that the syntax features can greatly enhance the prediction under cross-lingual transfer.
Particularly, GCN slightly outperforms TreeLSTM for encoding dependency tree.
Also, the pre-trained high-order features, as well as the XLM word representation are exceptionally helpful.
Finally, our \emph{BASIC} model can give the best-reported F1 score (i.e., 63.6\%), outperforming all the baseline SRL models.

\begin{table}[!t]
\begin{center}
  \caption{Results of cross-lingual transfer with multiple sources on argument role labeling using gold features. }
  \label{multi-source:ARL}
\resizebox{1.0\columnwidth}{!}{
  \begin{tabular}{lccccccc}
\hline
\multicolumn{1}{c}{Feature}  & ~EN~& ~DE~& ~FR~& ~IT~& ~ES~& ~PT~& ~FI~\\
\hline \hline
\multicolumn{8}{l}{$\bullet$ \textbf{Basic Word Feature (+Emb)}}\\ 
 Word & 42.4 & 36.2 & 43.8 & 37.5 & 30.9 & 36.8 & 27.4 \\
 \quad +Lemma & 43.8 & 38.7 & 47.6 & 39.6 & 35.8 & 39.6 & 28.9 \\
 \hline
\multicolumn{8}{l}{$\bullet$ \textbf{Auto Syntax Feature (+Emb)}}\\ 
\rotatebox{0}{POS} & 47.2 & 42.1 & 51.8 & 42.9 & 41.7 & 45.2 & 31.1 \\
\multicolumn{1}{l}{Dependency}&\multicolumn{6}{l}{}&\multicolumn{1}{l}{}\\
\multicolumn{1}{l}{\quad +TreeLSTM} &	53.4 & 50.6 & 60.1 & 51.7 & 47.0 & 49.8 & 37.3 \\
\multicolumn{1}{l}{\quad +GCN} & 54.0 & 52.4 & 61.1 & 52.8 & 49.5 & 50.1 & 37.5 \\
\cdashline{1-8}
Comb. &  \bf  60.3 &   \bf 58.0 &   \bf 64.9 &   \bf 60.1 &   \bf 54.5 &   \bf 54.2 &   \bf 43.2 \\
\hline
\multicolumn{8}{l}{$\bullet$ \textbf{Gold Syntax Feature (+Emb)}}\\ 
\rotatebox{0}{POS} & 48.3 & 46.3 & 54.4 & 43.9 & 43.0 & 46.8 & 31.7 \\
\multicolumn{1}{l}{Dependency: }&\multicolumn{6}{l}{}&\multicolumn{1}{l}{}\\
\multicolumn{1}{l}{\quad +TreeLSTM} &	55.2 & 53.1 & 62.1 & 54.0 & 51.9 & 50.6 & 38.2 \\
\multicolumn{1}{l}{\quad +GCN} &	56.0 & 55.3 & 64.7 & 55.2 & 52.4 & 54.0 & 40.1 \\
\cdashline{1-8}
Comb. &	  \bf 60.8 &   \bf 60.4 &  \bf  67.7 &   \bf 60.9 &   \bf 57.7 &  \bf  55.3 &   \bf 45.2 \\
\hline
\multicolumn{8}{l}{$\bullet$ \textbf{Pre-trained High-order Feature (+Comb. Gold Syn.)}}\\ 
\rotatebox{0}{HO.} & 63.2 & 62.4 & 70.5 & 62.7 & 59.4 & 58.3 & 47.3 \\
\hline
\multicolumn{8}{l}{$\bullet$ \textbf{Contextual Word Representation (+Comb. Gold Syn. +HO.)}}\\ 
\rotatebox{0}{ELMo} & 64.0 & 63.1 & 70.7 & 64.1 & 63.6 & 57.9 &  49.3   \\
\rotatebox{0}{BERT} & 65.9 & 64.2 & 72.2 & 65.5 &   \bf 64.2 &   \bf 59.7 & 49.8 \\
\rotatebox{0}{XLM} &   \bf 66.7 &   \bf 64.6 &  \bf  72.6 &  \bf  66.3 & 64.0 & 59.4 &   \bf 50.6 \\
\hline\hline
\multicolumn{8}{l}{$\bullet$ \textbf{SRL Model (+Comb. Gold Syn. +HO. +XLM)}}\\ 
\rotatebox{0}{He17-DeepAtt} & 64.6 & 62.6 & 72.0 & 64.9 & 62.6 & 58.2 & 49.8 \\
\rotatebox{0}{He18-Joint} & 64.8 & 62.9 & 72.1 & 63.2 & 63.5 & 58.7 & 50.1 \\
\hdashline
\emph{PGN}   &  \bf 68.3 &  \bf 67.4 & \bf  74.9 &  \bf 68.7 & \bf 67.5 &  \bf 63.3 & 53.1 \\
\emph{MoE} & 67.9 & 66.1 & 72.9 & 67.1 & 65.1 & 59.7 & 51.9 \\
\emph{MAN-MoE} & 68.2 & 66.4 & 73.0 & 67.9 &  66.3 & 60.6 & \bf 53.4 \\
\hline
\end{tabular}
}
\end{center}
\end{table}

\subsection{Bilingual Transfer from Each Source Language}

Further, in order to comprehensively study the usefulness of each types of cross-lingual features, we investigate the bilingual transfer learning from other sources,
where each source language is used as the source languages, respectively.
To make the performances more comparable, we mainly evaluate on the representative settings of each features, including 1) the end-to-end extraction with gold POS feature, 2) with gold dependency tree, 3) with combined syntax and XLM input, respectively, as in Table \ref{fine:bilingual1}, and 4) the argument role labeling with combined syntax, high-order feature and XLM input
as in Table \ref{fine:bilingual4}.
Generally, the results share a lot of consistent tendencies with the single-source transfer from English.

First of all, the dependency syntax is more useful than POS tags, and the XLM contextualized representation can further boost the final performances.
% , for both end-to-end scheme and the standalone argument role labeling.
Following, we look into the individual bilingual SRL transferring, and examine the performance of each source-target language pair,
aiming to uncover which language benefits a target the most and trying to answer whether all source languages are useful for a target language.
The languages belonging to one same family can benefit each other greatly,
bringing better improvements in the majority of cases (i.e., EN--DE, FR--IT--ES--PT).
The above finding coincides with the prior cross-lingual study \cite{Fei06295,wang-etal-2019-cross}.

\begin{figure}[!t]
\centering
\includegraphics[width=.99\columnwidth]{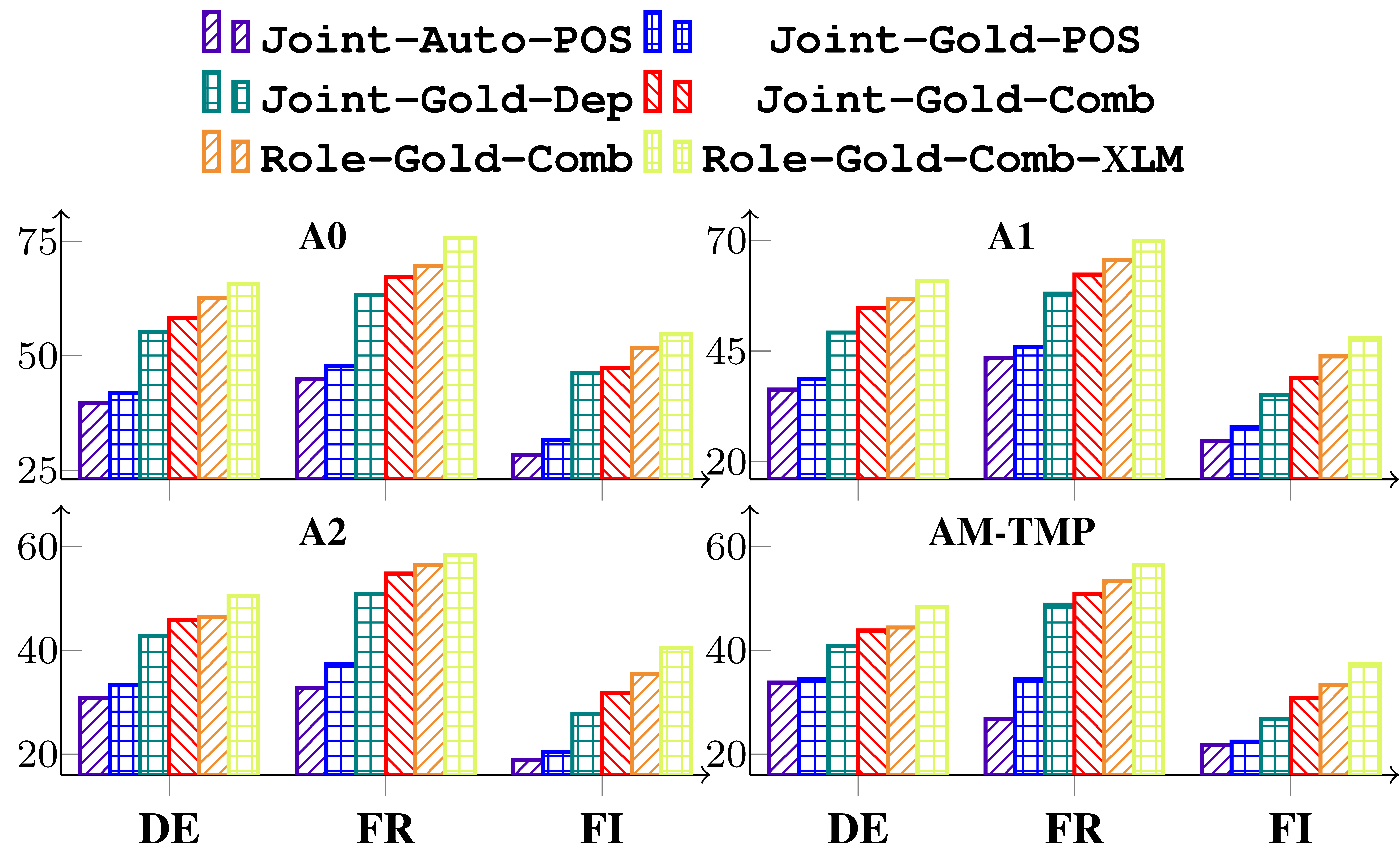}
\caption{
Performances of different argument labels.
}
\label{arg label}
\end{figure}

\subsection{Multi-Source Transfer}

We further explore the setting of multi-source transfer learning,
where all other languages except a given target language are used as the source languages.
Table \ref{multi-source:ARL} reports the results, based on the standalone argument role labeling, given predicates of sentences.
The overall performances share much similar pattern with the above single-source experiments, proving the differing usefulness of each types of cross-lingual features.
Most significantly, the cross-lingual transfer is more beneficial from multi-source transfer,
which can be seen between each correspondences in
Table \ref{fine:bilingual4} and Table \ref{Xling:En:ARL}, respectively.
% For example, the transfer to \emph{FR} from multi-source by \emph{PGN} model is 74.9\% F1 score, being higher than the counterpart from the best single-source (74.0\%, from \emph{IT}).
This further demonstrates the multi-source transfer advantage over the single-source transfer, because intuitively the former can offer more concentrated signals for the transfer.
Such finding is also consistent with previous studies \cite{Fei06295,lin-etal-2019-choosing}.

In addition, the linguistic-aware models can outperform those generic SRL models when multiple language inputs are used for training, such as the \emph{PGN} model, and the \emph{MoE} model.
Especially, our \emph{PGN} model is of the best utility, presenting the highest F1 scores across almost all the languages.
The underlying mechanism is that these linguistic-aware encoders can effectively make use of the language-specific information (i.e., $\bm{e}_{\mathcal{L}}$) as additional signals for better regulating the learning on multiple language inputs, such as, \emph{what to share and what to preserve privately}.
% The results prove the advances of the linguistic-aware encoders for capturing language-aware preferences for cross-lingual SRL.

\begin{figure}[!t]
\centering
\includegraphics[width=.92\columnwidth]{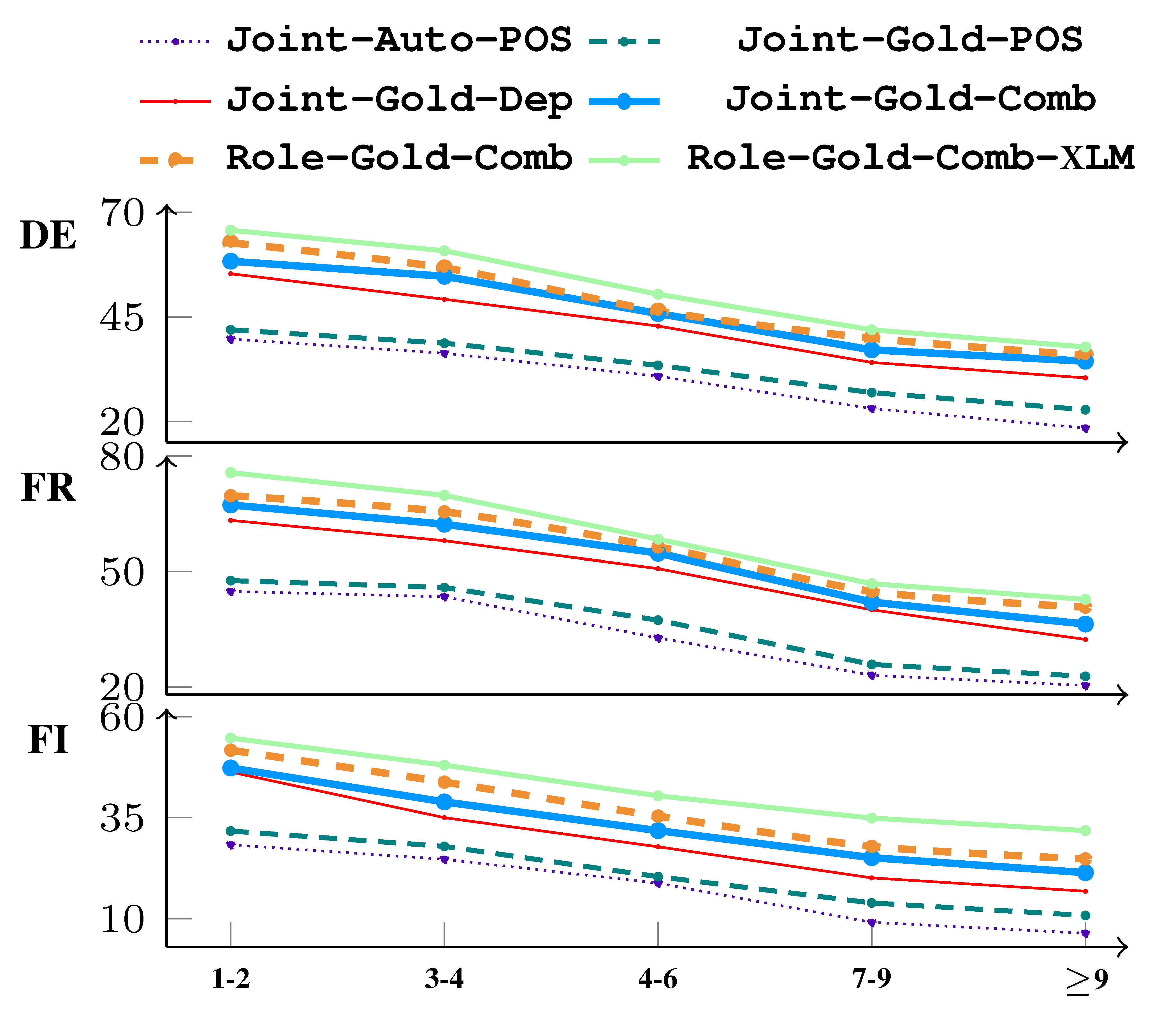}
\caption{
The effect of surface distances between predicates and arguments on SRL performances.
}
\label{prd distance}
\end{figure}

\section{Discussion}

We finally conduct detailed analysis to understand the gains from each types of cross-lingual features.
We select three languages for analysis, including German (DE), French (FR) and Portuguese (FI), one language for each family,
and compare six experiment settings,
including the joint extraction with auto-predicted POS tags, gold POS tags, gold dependency tree and combined gold syntax with high-order feature, and also the standalone argument role labeling with combined gold syntax and high-order feature with additional XLM representation.

\textbf{Performances on argument role labeling:}
% First, we investigate the cross-lingual SRL performances in terms of SRL Roles.
We select four representative roles for comparison,
including \texttt{A0} (\emph{Agent}), \texttt{A1} (\emph{Patient}), \texttt{A2} (\emph{Instrument, Benefactive, Attribute}) and \texttt{AM-TMP} (\emph{Temporal}),
and report their F1 scores.
Fig. \ref{arg label} shows the results.
As a whole, the role \texttt{A0} achieves the best F1 scores across all languages and all features,
\texttt{A1} ranks the second, and \texttt{A2} and \texttt{AM-TMP} are slightly worse.
The tendency could be accounted for by the distribution of these labels,
where \texttt{AM-TMP} have lower frequencies than \texttt{A0} and \texttt{A1}.

In addition, the tendencies across different models for all three languages and all labels
are identical, where the combined gold syntax helps best, and can be further improved by XLM representation.
The observation is consistent with the prior results,
demonstrating the stability and also further confirming the following observations.
1) The gold syntax feature is more useful than auto-predicted one;
2) The dependency structure makes more contribution for the SRL transfer;
3) Combined gold syntax with high-order feature together give improved helpfulness;
4) Contextualized word representations can enhance the performances.

\textbf{The effect of the distances between arguments and predicates:}
% Second, we study the SRL performances in terms of the distance to the predicate word.
Intuitively, long-distance relations are more difficult,
thus we expect that the SRL performance would decrease as the distance increases,
as SRL actually detects the relationship between the role words and their predicates.
Fig. \ref{prd distance} shows the F1 scores.
First, for all the settings we can see that the SRL performance drops by longer distances,
which confirms our intuition.
In addition, the tendency between different experiments is the same as the overall results, while those results with dependency tree features are better than that without it.
This demonstrates the effectiveness of dependency information on relieving the long-range dependency problem in the cross-lingual scenario.

\section{Conclusion}

This paper is dedicated to filling the gap of cross-lingual SRL with model transfer method.
Based on end-to-end SRL scheme of detecting all the predicates and labeling the argument roles with a linguistic-aware SRL model, we exploit widely employed cross-lingual features under a variety of settings.
We explore surface words, lemmas, POS tags and syntactic dependency trees, pre-trained high-order feature, as well as the contextualized multilingual word representations, including ELMo, BERT and XLM.
In addition, we compare the distinct contributions by gold standard syntactic features and auto-predicted features for the cross-lingual SRL, under the bilingual transferring where the one language is adopted as the source language, and the multi-source transferring where for each target language all the remaining six languages are used as the source languages, respectively.

Experiments are conducted based on Universal Proposition Bank corpus across seven languages.
Results show that the performances of cross-lingual SRL are varying by using different cross-lingual features, as well as whether the gold or auto features are applied.
The gold syntax feature is much more crucial for cross-lingual SRL, compared with the automatic features.
Besides, the universal dependency structure features offer the strongest utility, among all the syntax features, while combining all the syntax inputs, it gives the best help.
In addition, the pre-trained high order feature significantly contributes the cross-lingual transfer, providing rich semantics for the task, 
and the multilingual contextualized word representations greatly enhance the performances.
Last but not the least, linguistic-aware encoders are much crucial for the cross-lingual SRL.

\bibliographystyle{IEEEtran}
\bibliography{IEEEfull}

\end{document}